\documentclass[11pt]{article}

\usepackage{acl}

\usepackage{times}
\usepackage{latexsym}
\usepackage[T1]{fontenc}
\usepackage[utf8]{inputenc}
\usepackage{microtype}
\usepackage{inconsolata}
\usepackage{graphicx}
\usepackage{booktabs}
\usepackage{amsmath}
\usepackage{multirow}

\title{A Task-State Representation for Long-Horizon Mobile GUI Agents}

\author{
\textbf{Yujie Zheng}$^{2*}$,
\textbf{Zikang Liu}$^{1*}$,
\textbf{Xin Zhao}$^{1\dagger}$,
\textbf{Ji-Rong Wen}$^{1}$ \\
$^{1}$Gaoling School of Artificial Intelligence, Renmin University of China \\
$^{2}$School of Software, Beihang University \\
\texttt{\{janeyujie555,jasonlaw8121,batmanfly\}@gmail.com, jrwen@ruc.edu.cn} \\
}

\begin{document}

\maketitle

\begingroup
\renewcommand{\thefootnote}{\fnsymbol{footnote}}
\footnotetext[1]{Equal contribution.}
\footnotetext[2]{Corresponding author.}
\endgroup

\begin{abstract}
While long-horizon mobile GUI agents typically rely on thought-action-observation loops, they struggle to separate persistent task states from transient screen observations. As execution histories grow, this entanglement imposes a severe context burden, causing agents to forget initial requirements, hallucinate progress, or repeatedly interact with stale interfaces. To address this, we introduce Task-State Representation (TSR)—a training-free framework that explicitly decouples task state from sensory input. Acting as a lightweight external wrapper, TSR maintains three structured components: a global instruction summary, a dynamic progress tracker for subgoals, and a transition-aware action verifier. By continuously updating through pre- and post-action visual comparisons, TSR effectively guides the agent's reasoning without requiring architectural modifications. Experiments across four mobile GUI benchmarks validate TSR's effectiveness, yielding up to a 12 absolute point increase in success rate on complex cross-application and memory-intensive tasks.
\end{abstract}

\section{Introduction}

Automating mobile tasks via graphical user interfaces (GUIs) remains a long-standing goal in the development of intelligent agents.
Recent multimodal large language models (MLLMs)~\cite{liu2023visual} have facilitated prompt-based GUI actors that observe screenshots, reason about the current state, and generate executable actions, such as tapping, typing, or scrolling \citep{zhang2025appagent,hong2024cogagent}.
The dominant paradigm for handling long-horizon tasks adopts a thought-action-observation~\cite{yao2022react} loop: at each step, the actor receives the task instruction, a window of recent screenshots, and a history of previous reasoning and actions, and subsequently generates a new reasoning trace followed by an action \citep{zhang2024android,rawles2025androidworld}.
This append-all design relies on an implicit assumption---that the actor can reliably maintain awareness of the overall task goal and the cumulative progress from an ever-growing raw trajectory.

In practice, this assumption breaks down as the steps increases.
We identify three recurring failure modes in long-horizon mobile benchmarks:
(1)~\emph{goal drift}, where the actor gradually loses sight of the original task after observing numerous intermediate screens;
(2)~\emph{progress hallucination}, where the actor lose access to earlier visual observations and fabricates past states when reasoning about cumulative progress;
and (3)~\emph{stale-screen repetition}, where the actor misinterprets a delayed update in the user interface as a failed action and enters a localized loop.

These failures are not solely attributable to limitations in the model's capacity.
Rather, they arise from a structural deficiency in how the input of the actor is organized: the standard prompt conflates two fundamentally different categories of information---\emph{persistent task state} (the request of the user, the accomplished subgoals, and the remaining steps) and \emph{transient observation state} (the content displayed on the current screen).
Without an explicit mechanism to separate and maintain the former, the actor must re-derive the task progress from the raw history at every step---a burden that scales linearly with the length of the trajectory.

We propose a \textbf{task-state representation} that addresses this separation.
The representation is maintained externally to a fixed GUI actor and is updated at each step by a training-free state updater that compares the pre-action and post-action screenshots.
It comprises three functional views: a \emph{global task-state summary} that preserves the original instruction, a \emph{progress tracker} that records the completed and remaining subgoals, and a \emph{transition-aware focus} that verifies the effectiveness of the preceding action and guides the next decision.
The resulting state block is serialized and injected into the prompt of the actor, requiring no model retraining and architectural modification.

We evaluate two base models across four mobile GUI benchmarks: MobileWorld~\cite{kong2025mobileworld}, AndroidWorld~\cite{rawles2025androidworld}, MemGUI-Bench~\cite{liu2026memgui}, and VenusBench-Mobile~\cite{gong2026venusbench}. Our approach improves success rates in most configurations, with gains of up to 12\% on long-horizon tasks.
Our contributions are:

\begin{itemize}
    \item A task-state representation that separates persistent task state from transient observations for long-horizon mobile GUI agents.
    \item Empirical evaluation showing consistent improvements across four benchmarks and two base models, with ablations suggesting synergistic necessity of all three state views.
    \item Deeper analysis on the effect of structured task-state, identifying task horizon and state-tracking demand as key moderating factors.
\end{itemize}
\section{Method}

\begin{figure*}[t]
\centering
\includegraphics[width=\textwidth]{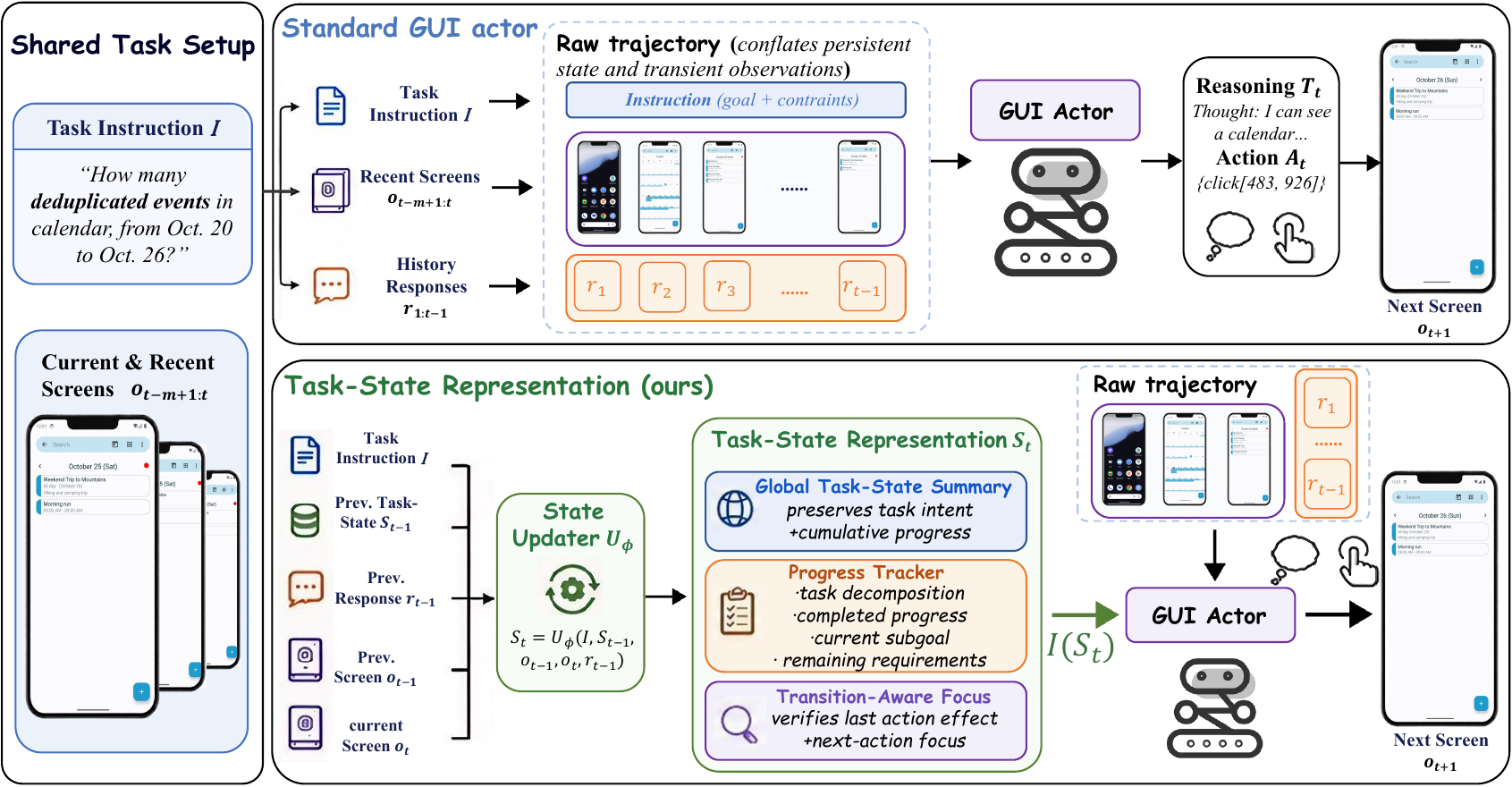}
\caption{Overview of the task-state representation. At each step, the state updater compares pre- and post-action screenshots and updates a structured state block which is then injected into the actor's input.}
\label{fig:framework}
\end{figure*}

\subsection{Problem Formulation}

We formulate mobile GUI automation as a sequential decision-making problem.
Given a task instruction $I$ and an initial screen observation $o_0$, a GUI actor $\pi_\theta$ repeatedly observes the current screen and generates a response $r_t = (T_t, a_t)$, where $T_t$ denotes a reasoning trace and $a_t$ represents an executable action drawn from a predefined action space $\mathcal{A}$ (detailed in Appendix~\ref{app:action-space}).
Following the execution of $a_t$, the environment transitions to a new screen state $o_{t+1}$, and the rollout continues until task completion or a predefined step limit.

To reduce context overhead, existing systems retain only the $m$ most recent screenshots alongside the full text history~\citep{li2026rethinking}.
Formally, at step $t$, the actor generates:
\[
a_t = \pi_\theta(I,\; O_t,\; H_{t-1}),
\]
where $O_t = \{o_{t-m+1}, \ldots, o_t\}$ denotes the recent observation window and $H_{t-1} = \{(T_1, a_1), \ldots, (T_{t-1}, a_{t-1})\}$ is the interaction history comprising all prior reasonings and actions.
This design preserves local visual continuity but discards earlier observations. Consequently, the actor must infer task progress from past responses, a growing burden in long trajectories that leads to the potential failures.

\subsection{Task-State Representation}

To address these limitations, we introduce an externally maintained task-state representation $S_t$, which is updated at each step and injected into the actor's input.
The state comprises three functional views, which are detailed below.
Rather than operating as separate modules, these views are fields within a unified state object and are updated jointly at each step.

\paragraph{Global Task-State Summary.}
When early visual contexts are truncated, the actor is prone to losing sight of the initial task goal. 
To counteract this intent decay, we design a summary module to record the persistent semantics of the task along with the cumulative progress.
The primary role of this summary is to maintain the visibility of the original task instruction, even when recent screenshots display only a narrow slice of the interface.
This mechanism prevents the actor from drifting away from the intent of the user during long sequences of interaction.

\paragraph{Progress Tracker.}
During complex long-horizon tasks, the agent often loses track of their overarching goals, creating severe ambiguity in evaluating current progress.
To resolve this ambiguity, we design a progress tracker that decomposes the task into atomic requirements and tracks the execution status of each.
It maintains four fields: task decomposition, completed milestones, current subgoal, and remaining requirements.
By making verified progress and unresolved steps explicit, the tracker mitigates the risk of the actor hallucinating earlier observations or terminating prematurely.

\paragraph{Transition-Aware Focus.}
Furthermore, since $H_{t-1}$ reflects past intentions rather than environmental feedback, the actor risks repetitive loops during system delays or failures. To close this loop, the transition view evaluates the previous action by comparing observations $o_{t-1}$ and $o_t$. If the outcome is uncertain, the representation prevents blind repetition by generating a \emph{next-action focus}—such as verifying the state or refreshing a list—to guide the actor's next decision.

\subsection{State Update and Actor Injection}

At the beginning of a task, the initial state $S_0$ is derived solely from the task instruction $I$.
At each subsequent step $t$, following the execution of action $a_{t-1}$ by the actor and the generation of observation $o_t$ by the environment, the state updater $\mathcal{U}_\phi$ receives the task instruction, the previous state, the previous response of the actor, and the screenshots captured before and after the action.
It then produces the updated state through a single function call:
\[
S_t = \mathcal{U}_\phi(I,\; S_{t-1},\; r_{t-1},\; o_{t-1},\; o_t).
\]
The updater $\mathcal{U}_\phi$ is implemented via a prompted LLM that outputs a structured JSON object encompassing all three views. Subsequently, the actor receives the standard context augmented with a serialized rendering of the task state:
\[
a_t = \pi_\theta(I,\; O_t,\; H_{t-1},\; \mathcal{I}(S_t)),
\]
where $\mathcal{I}(S_t)$ denotes an injection function that formats the three state views into a text block appended to the prompt of the actor.
Crucially, $\pi_\theta$ remains \emph{fixed} throughout, so that the entire mechanism operates externally at inference time without any training or architectural modifications.
\section{Experiments}

\subsection{Experimental Setup}

We evaluate our approach on four online GUI benchmarks, including MobileWorld \citep{kong2025mobileworld}, AndroidWorld~\citep{rawles2025androidworld}, MemGUI-Bench~\citep{liu2026memgui} and VenusBench-Mobile~\citep{gong2026venusbench}. We implement the standard GUI actor based on previous studies~\cite{kong2025mobileworld} as baseline. More details are presented in Appendix~\ref{app:exp-details}.

\subsection{Main Results}

Table~\ref{tab:main} reports success rates and average steps across all settings. We list our findings below.

\begin{table*}[t]
\centering
\small
\setlength{\tabcolsep}{10pt}
\begin{tabular}{l l | c c c c c}
\toprule
\multirow{2}{*}{Model} & \multirow{2}{*}{Benchmark} & Baseline & +Context & \multirow{2}{*}{$\Delta$ SR} & Baseline & +Context \\
 & & (SR) & (SR) & & (Avg steps) & (Avg steps) \\
\midrule
Qwen3.5-plus & MobileWorld & 43.00 & 55.00 & +12.00 & 25.1 & 27.0 \\
Qwen3.5-plus & AndroidWorld & 61.21 & 57.76 & -3.45 & 12.8 & 13.9 \\
Qwen3.5-plus & MemGUI-Bench & 26.56 & 28.91 & +2.35 & 60.5 & 59.8 \\
Qwen3.5-plus & MemGUI-Memory & 23.48 & 28.70 & +5.22 & 65.4 & 64.5 \\
Qwen3.5-plus & VenusBench-Mobile & 15.25 & 20.34 & +5.09 & 38.7 & 38.2 \\
\midrule
Kimi-k2.5 & MobileWorld & 49.00 & 58.00 & +9.00 & 29.6 & 28.3 \\
Kimi-k2.5 & AndroidWorld & 51.72 & 55.17 & +3.45 & 15.5 & 15.8 \\
Kimi-k2.5 & MemGUI-Bench & 42.19 & 42.97 & +0.78 & 51.6 & 55.8 \\
Kimi-k2.5 & MemGUI-Memory & 37.39 & 40.87 & +3.48 & 55.5 & 60.6 \\
Kimi-k2.5 & VenusBench-Mobile & 22.03 & 26.27 & +4.24 & 41.2 & 38.3 \\
\bottomrule
\end{tabular}
\caption{Performance comparison of baseline and our proposed task-state representation.}
\label{tab:main}
\end{table*}

First, the most pronounced improvements appear on MobileWorld (+12\% for Qwen3.5-plus, +9\% for Kimi-k2.5), the benchmark with the longest average trajectories and cross-application dependencies.
This is consistent with the hypothesis that explicit progress tracking is most beneficial when the agent navigate extended interaction sequences.

Besides, on memory-intensive and user-centric benchmarks~(MemGUI-Memory, VenusBench-Mobile), both models benefit (+3.48\% to +5.22\%).
However, the representation is not universally positive: Qwen3.5-plus degrades by 3.45\% on AndroidWorld while Kimi-k2.5 improves by 3.45\% on the same benchmark, suggesting that the utility of structured task-state tracking depends on the base model's inherent planning capability and the task complexity distribution.

Third, higher success rates do not always require more steps.
In several cases (Kimi-k2.5 on VenusBench-Mobile: 41.2$\to$38.3 steps; Qwen3.5-plus on MemGUI-Memory: 65.4$\to$64.5 steps), the representation enables more direct task completion.
Conversely, on MobileWorld the step count increases alongside SR, indicating that the representation helps agents persist through complex tasks they would otherwise abandon.

\subsection{Ablation Study}

We ablate on Qwen3.5-plus by separately removing the task-state summary, progress tracker, and transition-aware focus from the representation while still updating the internal state.
Results are shown in Table~\ref{tab:ablation}.

\begin{table}[t]
\centering
\small
\begin{tabular}{l c c}
\toprule
Method & MobileWorld & AndroidWorld \\
\midrule
Baseline & 43.00 & 61.21 \\
\midrule
+Context w/o summary & 50.00 & 59.48 \\
+Context w/o tracker & 49.00 & 61.21 \\
+Context w/o transition & 48.00 & \textbf{63.79} \\
+Context full & \textbf{55.00} & 57.76 \\
\bottomrule
\end{tabular}
\caption{Ablation on Qwen3.5-plus (Success Rate, \%).}
\label{tab:ablation}
\end{table}

Firstly, we discover that on MobileWorld, the full representation outperforms every ablated variant by 5--7\%, suggesting that the three components function synergistically.
Among individual removals, dropping the transition-aware focus causes the largest degradation (55$\to$48), suggesting that verifying action effectiveness is the most critical signal for long-horizon tasks. However, the pattern reverses on AndroidWorld. Specifically, removing the transition module improves over the baseline (63.79 vs.\ 61.21), while the full representation hurts.
This indicates that for shorter, single-application tasks, transition verification introduces unnecessary decision noise.
The task-state representation is therefore best viewed as conditionally beneficial---most valuable when tasks demand persistent state tracking across many steps.

\subsection{Case Study}

\begin{figure*}[t]
\centering
\includegraphics[width=\textwidth]{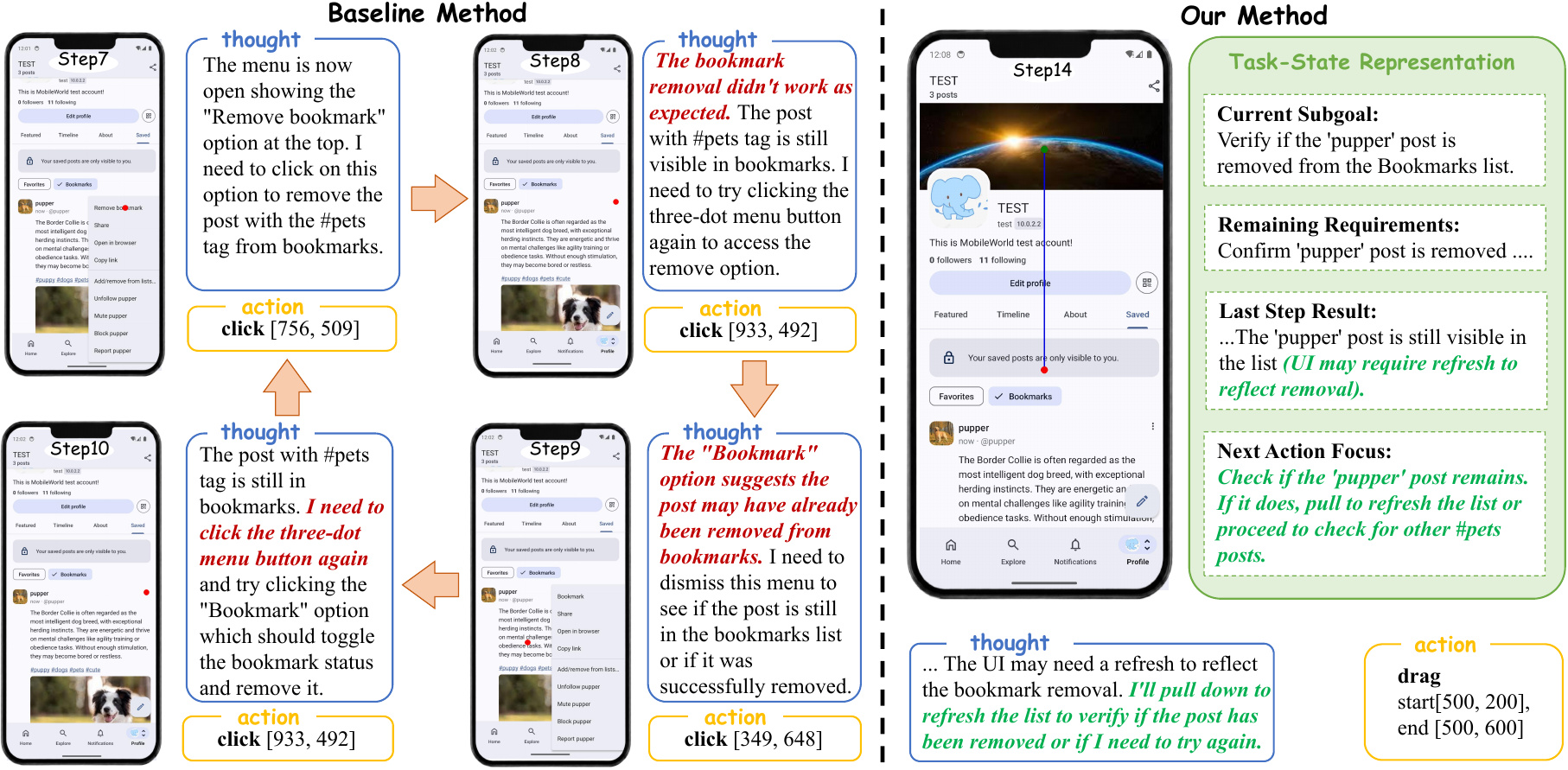}
\caption{Stale-screen recovery. The baseline repeats the same removal action against a delayed UI update. The task-state representation detects the uncertain transition and redirects the actor to verify before retrying.}
\label{fig:case-stale}
\end{figure*}

\begin{figure*}[t]
\centering
\includegraphics[width=\textwidth]{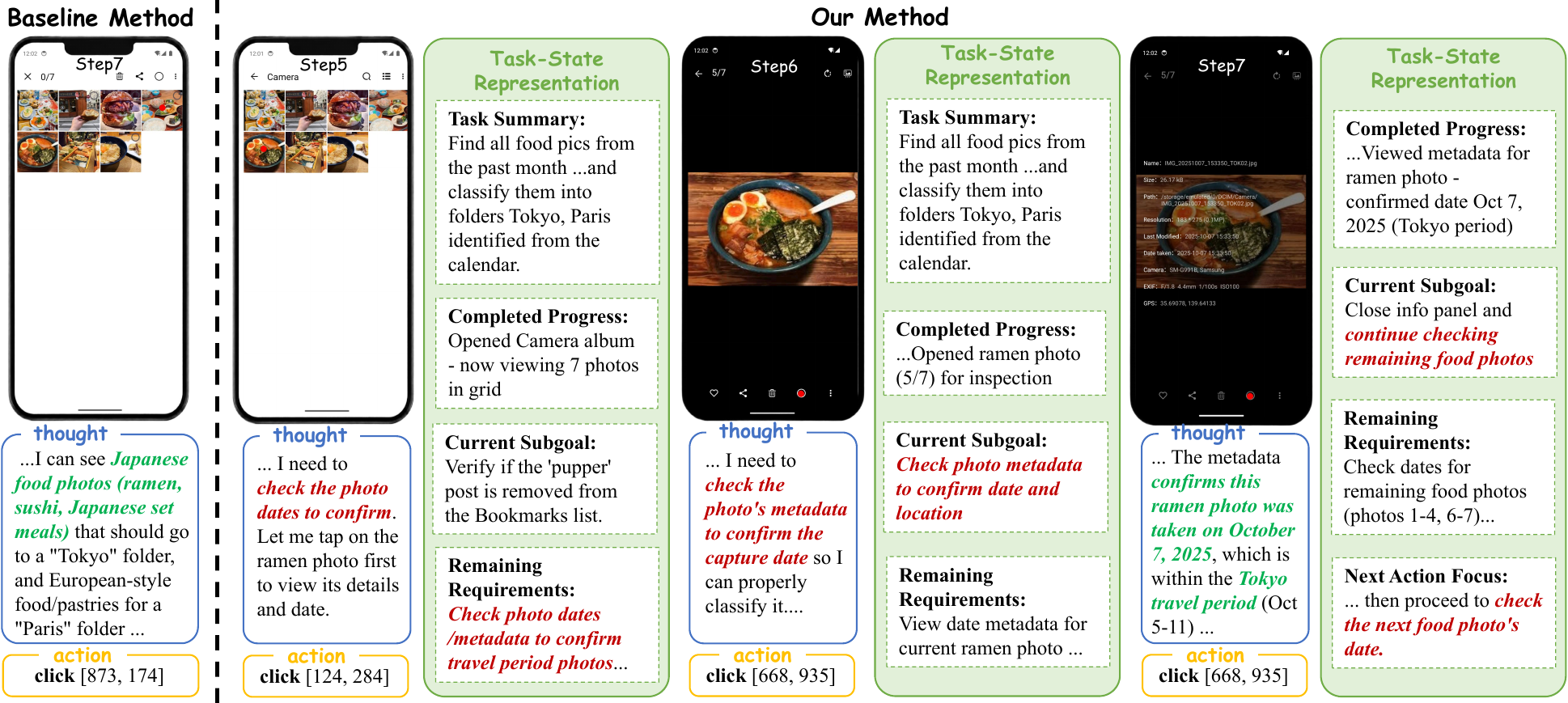}
\caption{Over-decomposition failure. The state updater decomposes the task too aggressively, causing the actor to inspect items individually rather than using global visual inference, exceeding the step budget.}
\label{fig:case-overdecomp}
\end{figure*}

Figure~\ref{fig:case-stale} illustrates a bookmark-management task where the baseline enters a repetition loop after a delayed UI update; the transition-aware focus detects the uncertain effect and guides the actor to verify before retrying.
Figure~\ref{fig:case-overdecomp} shows a failure mode: when the updater decomposes a task too aggressively, the actor inspects items individually instead of leveraging global visual inference, exceeding the step budget.
The task-state representation should therefore be treated as guidance rather than an override of current-screen evidence.
\section{Related Work}

\paragraph{Mobile GUI Agents.}
Recent multimodal LLMs~\cite{liu2023visual,liu2025less} have enabled prompt-based GUI actors that translate instructions into touch actions on mobile screens \citep{zhang2025appagent,hong2024cogagent}.
Subsequent work improves grounding accuracy \citep{cheng2024seeclick} and scales toward long-horizon tasks via hierarchical planning \citep{lu2026ui}, history compression \citep{li2026rethinking} or active loading~\cite{liu2025pal}.
However, existing methods often require retraining, limiting their use as general wrappers. To address this, we propose a training-free task-state representation that enriches prompts without altering the base agent.

\paragraph{Long-Horizon Context for LLM Agents.}
Managing growing interaction trajectories is a central challenge for LLM agents.
Existing approaches primarily address this through trajectory compression: HiAgent \citep{hu2025hiagent} chunks working memory by subgoals, IterResearch \citep{chen2025iterresearch} periodically reconstructs a condensed workspace, and summarization-based methods learn to shorten history via RL \citep{lu2025scaling}.
These methods treat the trajectory as a monolithic stream to be shortened; in contrast, we maintain a separate structured representation alongside the actor's unmodified observation history.

\section{Conclusion}

In this paper, we propose a task-state representation that decouples persistent task state from transient screen observations to counter goal drift, progress hallucination, and stale-screen repetition for long-horizon mobile GUI agents. Our approach augments a fixed actor with three externally maintained views: a global task-state summary, a progress tracker, and a transition-aware focus, without any architectural modification. Evaluated across four benchmarks, our method improves success rates by up to 12 \% on long-horizon cross-app tasks and yields consistent 3–5 \% gains on memory-intensive tasks, validating the effectiveness of our approach.

\section*{Limitations}

The state updater can propagate errors: if it incorrectly marks a subgoal as completed or omits a remaining requirement, the actor conditions subsequent decisions on a flawed state, potentially compounding mistakes across steps.
Additionally, the updater introduces one extra LLM call per step (1024 tokens), increasing both latency and cost; for short tasks where the baseline already succeeds, this overhead provides no benefit and may degrade performance through over-decomposition.
All experiments are conducted on Android-based mobile benchmarks; whether the representation generalizes to desktop or web GUI environments remains untested.
Finally, the current design updates all three state views jointly---a more selective mechanism that activates views only when needed could reduce noise on simpler tasks.

\bibliography{custom}

\newpage
\appendix

\section{Experimental Details}
\label{app:exp-details}

\paragraph{Datasets.}
We evaluate on four benchmarks spanning different difficulty axes.
\textbf{MobileWorld} \citep{kong2025mobileworld}: long-horizon, cross-application workflows; we exclude agent-user interaction and MCP-augmented tasks, retaining 100 GUI-only tasks.
\textbf{AndroidWorld} \citep{rawles2025androidworld}: reproducible Android tasks with programmatic reward verification.
\textbf{MemGUI-Bench} \citep{liu2026memgui}: 115 memory-intensive tasks requiring cross-temporal and cross-spatial recall (denoted MemGUI-Memory for the memory-specific subset).
\textbf{VenusBench-Mobile} \citep{gong2026venusbench}: realistic user-centric tasks with capability diagnostics; we exclude unsupported tasks, leaving 118 tasks.

\paragraph{Baselines.}
Our baseline is the standard MobileWorld actor, which instantiates $\pi_\theta$ as defined in Section~2.1.
We compare this baseline against the same actor augmented with the task-state representation $S_t$.

\paragraph{Implementation Details.}
We use two multimodal LLMs as the underlying model for both the actor and the state updater: Qwen3.5-plus \citep{yang2025qwen3} and Kimi-k2.5 \citep{team2026kimi}, accessed via Bailian API with temperature 0.

\section{Action Space Definition}
\label{app:action-space}

The action space $\mathcal{A}$ differs between benchmarks. Table~\ref{tab:action-space} lists all available actions.

\begin{table}[ht]
\centering
\small
\begin{tabular}{lll}
\toprule
Action & Parameters & Scope \\
\midrule
\texttt{click} & coordinate $[x,y]$ & All \\
\texttt{double\_tap} & coordinate $[x,y]$ & All \\
\texttt{long\_press} & coordinate $[x,y]$ & All \\
\texttt{drag} & start, end coords & All \\
\texttt{input\_text} & text string & All \\
\texttt{answer} & text string & All \\
\texttt{navigate\_home} & --- & All \\
\texttt{navigate\_back} & --- & All \\
\texttt{scroll} & direction & All \\
\texttt{status} & complete/infeasible & All \\
\texttt{wait} & --- & All \\
\texttt{keyboard\_enter} & --- & All \\
\texttt{open\_app} & app name & AW \\
\texttt{swipe} & direction & AW \\
\bottomrule
\end{tabular}
\caption{Action space $\mathcal{A}$. AW = AndroidWorld only. Coordinates are normalized to $[0, S]$ where $S$ is a model-specific scale factor (1000 for Qwen3.5-plus, 1 for Kimi-k2.5). \texttt{scroll} moves content; \texttt{swipe} performs system-level finger gestures. \texttt{answer} and \texttt{status} terminate the episode.}
\label{tab:action-space}
\end{table}

\section{LLM Configuration}
\label{app:llm-config}

\begin{table}[ht]
\centering
\small
\setlength{\tabcolsep}{3pt}
\begin{tabular}{@{}lcc@{}}
\toprule
Parameter & Actor $\pi_\theta$ & Updater $\mathcal{U}_\phi$ \\
\midrule
Temperature & 0.0 & 0.0 \\
Max output tokens & 2048 & 1024 \\
Obs.\ window $m$ & 3 & 2 (pre + post) \\
\midrule
\multicolumn{3}{@{}l}{\textit{Model-specific}} \\
\midrule
Scale $S$ (Qwen3.5-plus) & \multicolumn{2}{c}{1000} \\
Scale $S$ (Kimi-k2.5) & \multicolumn{2}{c}{1} \\
API & \multicolumn{2}{c}{Bailian} \\
\bottomrule
\end{tabular}
\caption{Representation and inference configurations. Both components use the same underlying model (Qwen3.5-plus \citep{yang2025qwen3} or Kimi-k2.5 \citep{team2026kimi}) accessed via provider APIs.}
\label{tab:llm-config}
\end{table}

The maximum number of interaction steps per task is set to 50 for MobileWorld; for AndroidWorld, MemGUI-Bench, and VenusBench-Mobile, we use the benchmark-specific preset limits defined by each environment.
The state updater is invoked at every step regardless of the benchmark.

\section{Actor Prompt Template}
\label{app:actor-prompt}

The actor receives a structured prompt comprising a role definition, action framework, execution principles, the task instruction, and the injected task-state representation $\mathcal{I}(S_t)$. The template below shows the MobileWorld variant; AndroidWorld omits \texttt{open\_app} and \texttt{swipe}. Coordinates are normalized to $[0, S]$ where $S$ is the model-specific scale factor.

\begin{small}
\begin{verbatim}
# Role: Android Phone Operator AI
You are an AI that controls an Android
phone to complete user requests.

# Action Framework
Respond with EXACT JSON format for one
of these actions:
| Action       | JSON Format             |
|--------------|-------------------------|
| click        | {"action_type":"click", |
|              |  "coordinate":[x,y]}   |
| long_press   | {"action_type":         |
|              |  "long_press",          |
|              |  "coordinate":[x,y]}   |
| drag         | {"action_type":"drag",  |
|              |  "start_coordinate":    |
|              |  [x1,y1],              |
|              |  "end_coordinate":      |
|              |  [x2,y2]}              |
| input_text   | {"action_type":         |
|              |  "input_text",          |
|              |  "text":"content"}     |
| answer       | {"action_type":"answer",|
|              |  "text":"response"}    |
| scroll       | {"action_type":"scroll",|
|              |  "direction":"down"}   |
| open_app     | {"action_type":         |
|              |  "open_app",            |
|              |  "app_name":"Calendar"} |
| status       | {"action_type":"status",|
|              |  "goal_status":         |
|              |  "complete"}            |

# Execution Principles
1. Communication: ALWAYS use 'answer'
   to reply to user questions.
2. Efficiency: Choose the simplest path.
3. Navigation: scroll = content move;
   swipe = system gesture.
4. Text input: Click input box first.

# Task Instruction
{I}

# Task-State Representation
## Global Summary
{cumulative task state}

## Progress Tracker
Task Decomposition:
- {decomposition items}
Completed Progress:
- {completed items}
Current Subgoal: {current subgoal}
Remaining Requirements:
- {remaining items}

## Transition-Aware Focus
Last Step Result:
- Effectiveness: {effective/uncertain}
- Outcome: {last_step_result}
Next Action Focus: {next_action_focus}

# Output Format
Thought: [Your analysis of current state]
Action: [Single JSON action]
\end{verbatim}
\end{small}

\section{State Updater Prompt Template}
\label{app:updater-prompt}

The state updater $\mathcal{U}_\phi$ receives the following prompt along with before/after screenshots ($o_{t-1}$, $o_t$) as image inputs.

\paragraph{System message.}
\begin{small}
\begin{verbatim}
You maintain a structured task-state
representation for a mobile GUI actor.
Return strict JSON only.
\end{verbatim}
\end{small}

\paragraph{User message.}
\begin{small}
\begin{verbatim}
Task Instruction: {I}

Previous Task-State Representation (JSON):
{S_{t-1}}

Last Step:
- Actor reasoning: {T_{t-1}}
- Action executed: {a_{t-1}}
- Screenshot before action: [image o_{t-1}]
- Screenshot after action: [image o_t]

Instructions:
1. Compare the before/after screenshots
   to determine action effectiveness.
2. Update task decomposition if the task
   requires multiple steps.
3. Preserve all verified completed
   progress from previous state.
4. Update the current subgoal.
5. Align remaining requirements.
6. Write next_action_focus to guide
   the actor's next decision.

Rules:
- Treat the previous state as persistent
  working memory across steps.
- Base judgments only on visible UI state;
  do not invent hidden information.
- Keep each field concrete and relevant.
- Return JSON only, no explanation.

Output JSON schema:
{
  "action_effective": true|false|null,
  "task_summary": "...",
  "task_decomposition": ["..."],
  "completed_progress": ["..."],
  "current_subgoal": "...",
  "remaining_requirements": ["..."],
  "last_step_result": "...",
  "next_action_focus": "..."
}
\end{verbatim}
\end{small}

\end{document}